\documentclass[lettersize,journal]{IEEEtran}
\usepackage{amsmath,amsfonts}
\usepackage{algorithmic}
\usepackage{algorithm}
\usepackage{array}
\usepackage[caption=false,font=normalsize,labelfont=sf,textfont=sf]{subfig}
\usepackage{textcomp}
\usepackage{stfloats}
\usepackage{url}
\usepackage{verbatim}
\usepackage{graphicx}
\usepackage{pifont}
\usepackage{cite}
\usepackage{makecell}
\usepackage{diagbox}
\usepackage{multirow}
\usepackage{amsthm}
\usepackage{colortbl}
\usepackage{xcolor}
\usepackage{color}
\newcommand{\etal}{\textit{et al.}}

\definecolor{SkyBlue}{RGB}{30,144,255}

\begin{document}

\title{ExpLLM: Towards Chain of Thought for Facial Expression Recognition}


\renewcommand{\thefootnote}{\fnsymbol{footnote}}

\author{Xing Lan\textsuperscript{1,4}, Jian Xue\textsuperscript{1}, Ji Qi\textsuperscript{2,4}, Dongmei Jiang\textsuperscript{3}, Ke Lu\textsuperscript{1,3} and Tat-Seng Chua\textsuperscript{4}
\\\textsuperscript{1}School of Engineering, University of Chinese Academy of Sciences
\\\textsuperscript{2}Department of Computer Science and Technology, Tsinghua University
\\\textsuperscript{3}Pengcheng Laboratory
~~\textsuperscript{4}School of Computing, National University of Singapore
\\
{\tt\small\url{https://starhiking.github.io/ExpLLM_Page}}

% \thanks{
% This work was supported by the National Natural Science Foundation of China (62236006, 62320106007, 62032022) 
% and the Scientific Research Program of the Beijing Municipal Education Commission (KZ201911417048). 
% (\emph{Corresponding author: Ke Lu})}%
% \thanks{Xing Lan, Jian Xue and Ke Lu are with University of Chinese Academy of Sciences, Beijing, China. 
% (email: lanxing19@mails.ucas.ac.cn; xuejian@ucas.ac.cn; luk@ucas.ac.cn).
% ~
% Ji Qi is with Tsinghua University, Beijing, China. (email: qijimrc@gmail.com).
% Dongmei Jiang is with Peng Cheng Laboratory, Shenzhen, China. (email: Jiangdm@pcl.ac.cn).
% Tat-Seng Chua is with National University of Singapore, Singapore. (email: chuats@comp.nus.edu.sg).
% }% <-this % stops a space
\thanks{
Under Review. 
}
}

\markboth{Journal of \LaTeX\ Class Files,~Vol.~14, No.~8, August~2021}%
{Shell \MakeLowercase{\textit{et al.}}: A Sample Article Using IEEEtran.cls for IEEE Journals}


\maketitle
\footnotetext[1]{Work done during visiting at NUS.}
\renewcommand{\thefootnote}{\arabic{footnote}}

\begin{abstract}
  
  % Facial expression recognition (FER) is a important task in multi media with profound implications across multiple domains.
  % However, analyzing the face details in expression is also important in recognize the facial expression.
  % At present, most work, like based on facial action units, just give the action unit and degree, no the interaction or relationship with the expression.
  % In this paper, we propose a novel method, called ExpLLM, which can generate accurate chain of thought for facial expression recognition.
  % We design a novel CoT, contains three perspectives: key observations, overall emotional interpretation and conclusion.
  % The key observations contains AU name and its intensity, as well as its potential emotions.
  % The overall emotional interpretation provide the analysis based on multiple AUs and their emotions, and point out the major and interactions with others.
  % The conclusion will present the final expression label based on the above analysis.
  % We design a novel Exp-CoT Engine to construct such expression CoT, and make the instruction-description data for training ExpLLM.
  % Extensive experiments on RAF-DB and AffectNet demonstrate that ExpLLM outperforms state-of-the-art methods in FER.
  % And generating the CoT of FER is benefical to the final accuracy.
  % ExpLLM also performs better than the newest GPT-4o in expression CoT generation, especially in the micro emotion which GPT-4o always fails.

Facial expression recognition (FER) is a critical task in multimedia with significant implications across various domains. 
However, analyzing the causes of facial expressions is essential for accurately recognizing them. 
Current approaches, such as those based on facial action units (AUs), typically provide AU names and intensities but lack insight into the interactions and relationships between AUs and the overall expression.
In this paper, we propose a novel method called ExpLLM, which leverages large language models to generate an accurate chain of thought (CoT) for facial expression recognition. 
Specifically, we have designed the CoT mechanism from three key perspectives: key observations, overall emotional interpretation, and conclusion. 
The key observations describe the AU's name, intensity, and associated emotions.
The overall emotional interpretation provides an analysis based on multiple AUs and their interactions, identifying the dominant emotions and their relationships. 
Finally, the conclusion presents the final expression label derived from the preceding analysis.
Furthermore, we also introduce the Exp-CoT Engine, designed to construct this expression CoT and generate instruction-description data for training our ExpLLM. 
Extensive experiments on the RAF-DB and AffectNet datasets demonstrate that ExpLLM outperforms current state-of-the-art FER methods. 
ExpLLM also surpasses the latest GPT-4o in expression CoT generation, particularly in recognizing micro-expressions where GPT-4o frequently fails.
% Additionally, generating a CoT for FER improves final accuracy. 

\end{abstract}


\begin{figure}[tbh]
  \begin{center}
  \includegraphics[width=\linewidth]{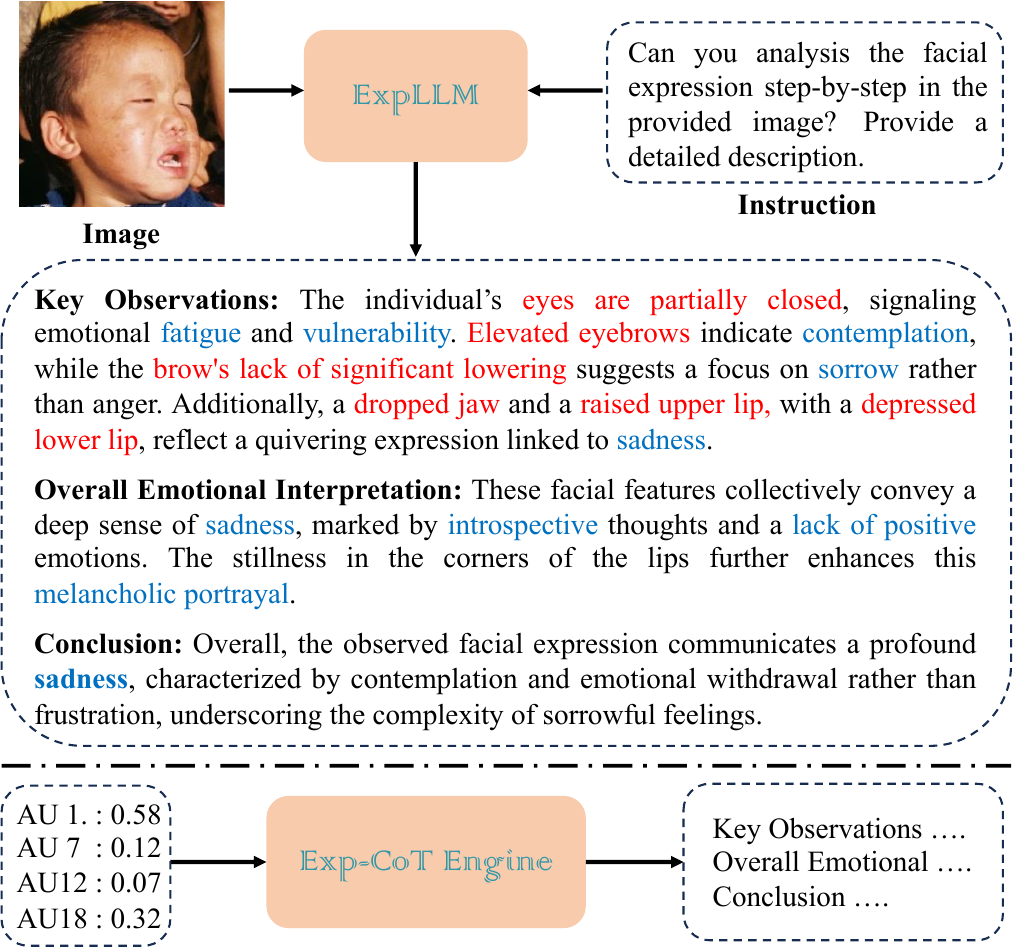}
  \end{center}
  \caption{
    The overall structure of the proposed methods.
    The upper part illustrates the ExpLLM. 
    By inputting both the instruction and the facial image into the ExpLLM, it sequentially generates the Expression CoT. 
    \textcolor{SkyBlue}{Blue text} represents emotional expressions, whereas \textcolor{red}{red text} denotes facial action units and their corresponding degrees.
    The lower part features the Exp-CoT Engine, which is designed to construct the CoT from three perspectives: key observations, overall emotional interpretation, and conclusion. 
    The Exp-CoT Engine utilizes the AU results to generate the expression CoT.
   }
  \label{fig: abstract}
  \end{figure}

\section{Introduction}
% 翻一下之前的 ppt，找一下 motivation

%  为什么这么设计， AU 需要过滤，找到合适的组合，以及哪个AU 起决定性因素。
%  因为根据 AU 推导出的表情是很复杂的，除了主要表情外，还会有更深层次的表情。
\IEEEPARstart{F}{acial} expressions~\cite{de2011facial,ekman1993facial}, which are a sophisticated blend of conscious reactions to specific stimuli, play a pivotal role in human communication. 
They allow individuals to express a wide range of emotions and intentions non-verbally, thereby enhancing interpersonal interactions and fostering social cohesion.
Facial expression recognition (FER) \cite{li2020deep,bettadapura2012face} has garnered significant attention in recent years due to its profound implications across multiple domains, 
ranging from psychology and human-computer interaction~\cite{shi2020human} to security~\cite{sivaram2019biometric} and healthcare~\cite{lee2021development}. 
For example, FER is employed in mental health assessments to gauge patients' emotional states~\cite{lee2021development}.
%  and in human-computer interaction~\cite{shi2020human} to enhance user experiences by enabling machines to respond to emotional cues. 
% Additionally, FER is employed in security for threat detection and in marketing to analyze consumer reactions, optimizing advertising strategies and product designs.

The Facial Action Coding System (FACS), introduced by Ekman \cite{ekman1978facial}, systematically encodes facial behavior through distinct movements of facial muscles, referred to as Action Units (AUs). 
In contrast to general facial expressions, action units provide a more intricate and detailed comprehension of human facial behavior~\cite{zhi2020comprehensive}. 
This is attributable to the fact that multiple AUs can manifest concurrently at different intensities, whereas FER typically involves single-label classification. 
Consequently, FACS-based FER methodologies have garnered widespread popularity and achieved notable success by offering comprehensive insights into facial expressions.

However, existing works still fall short of delivering an intuitive and transparent understanding of facial expressions. 
Various strategies have been made to integrate AUs into FER systems by either directly incorporating AU predictions into the network or utilizing AU prediction as an auxiliary task to provide the network with prior knowledge.
However, these approaches are not friendly to the analysis and explanation of facial expressions, as they fail to deliver a lucid and interpretable reasoning process for FER.
Additionally, the diverse and non-unique combinations of AUs complicate the process of deriving facial expressions from AUs.
These constraints hinder the practical application of FACS, such as offering nuanced feedback to therapists regarding a patient's emotional subtleties in psychological settings.

To mitigate these shortcomings, we are inspired by the recent success of multi-modal large language models (MLLMs). 
These models~\cite{wang2024exploring} have demonstrated a remarkable ability to reason over intricate and complex visual cues through instruction tuning following large-scale pre-training. 
In practice, MLLMs convert the discriminative tasks (such as regression~\cite{wang2024locllm}, classification~\cite{naeem2023i2mvformer}, etc.) into the text generative task based on large language models (LLMs). 
MLLMs have exhibited exceptional performance across various visual understanding tasks, including visual question answering, captioning, grounding, etc.

To this end, we propose an innovative model, termed \textbf{Exp}ression \textbf{L}arge \textbf{L}anguage \textbf{M}odel (ExpLLM), 
which can provide an intuitive and step-by-step analysis of facial expressions, as illustrated in Figure~\ref{fig: abstract}.
Meanwhile, we design a novel reasoning Chain-of-Thought (CoT) for FER, 
encompassing three perspectives: key observations, comprehensive emotional interpretation, and final conclusion.

Key observations include the AU name, its intensity, and the potential emotions it signifies.
Overall emotional interpretation considers multiple AUs and their interactions, identifying the predominant emotions and their relationships. 
The final conclusion presents the ultimate expression label based on the preceding analysis.
 
To facilitate the instruction tuning of ExpLLM, we have designed an innovative data engine named Exp-CoT Engine, which generates instruction-description pairs for ExpLLM’s instruction learning. 
Specifically, Exp-CoT Engine contains a AU model to capture the micro facial cues and utilizes ChatGPT to translate these cues into description.
Additionally, it includes the correctness checking mechanism to enhance the understanding and reasoning precision, and adopts the CoT format to standardize the descriptions, thereby enhancing the model's learning efficiency.

ExpLLm is able to simultaneously facilitate the generation of expression CoT and the recognition of expression labels. 
Our approach involves structuring the instruction-description and instruction-label data as multi-round dialogues. 
This iterative conversational framework enables the model to progressively enhance its comprehension of facial expressions, thereby augmenting its effectiveness in facial expression recognition tasks.

Extensive experimentation on FER datasets demonstrates that ExpLLM surpasses state-of-the-art methodologies in accuracy. 
Specifically, ExpLLM achieves accuracies of 91.03\% on RAF-DB~\cite{li2017reliable}, 65.93\% on AffectNet-7, and 62.86\% on AffectNet-8~\cite{mollahosseini2017affectnet}. 
Furthermore, cross evaluations are conducted on ExpW~\cite{zhang2015learning} and AffectNet-7, and the results indicate that ExpLLM outperforms those methods in FER tasks.
Additionally, we assess the quality of the generated expression descriptions, and the results indicate that ExpLLM produce accurate and coherent expression descriptions.
ExpLLM surpasses the latest GPT-4o in all prospective evaluations, particularly in recognizing micro cues where GPT-4o frequently fails.

In summary, the main contributions of this work are as follows:

\begin{itemize}
  \item An innovative model, named ExpLLM, provides an intuitive and step-by-step analysis of facial expressions using a novel reasoning Chain-of-Thought approach, enhancing transparency and interpretability in FER.

  \item A data construct engine, called Exp-CoT Engine, is designed to facilitate ExpLLM’s instruction tuning by generating instruction-description pairs. Exp-CoT Engine is equipped with advanced reasoning chains and detailed facial cue-capturing capabilities, ensuring robust and accurate expression analysis.
  
  \item Extensive experiments on FER datasets demonstrate that ExpLLM outperforms state-of-the-art approaches in accuracy. Meanwhile, ExpLLM can generate precise expression descriptions, surpassing the latest GPT-4o in all prospective evaluations.

\end{itemize}

% The remainder of this paper is organized as follows.
% Section~\ref{sec:relatedwork} reviews the related work.
% Section~\ref{sec:method} presents the proposed method.
% Section~\ref{sec:experiments} reports the experimental results.
% Finally, Section~\ref{sec:conclusion} concludes the paper.

\section{Related Work}
\label{sec:relatedwork}

In this section, we review the existing literature on Facial Expression Recognition, Facial Action Unit, and Multimodal Large Language Models. 
We first discuss traditional FER models and vision-language-based FER models. We then review recent advancements in AU detection, followed by an overview of the current landscape of MLLMs.
In the last, we introduce the concept of Chain of Thought (CoT) and its potential applications in FER.

\subsection{Facial Expression Recognition}
\textbf{Traditional FER Models:} 
Traditional Facial Expression Recognition models typically rely on a feature extractor followed by a classifier. 
Over the past decade, research on FER has primarily focused on overcoming real-world challenges such as occlusion \cite{zhao2021learning}, pose variation \cite{wang2019identity}, domain shift \cite{chen2021cross}, and label noise \cite{wang2020suppressing,zhao2021robust,wu2023net}. 
For example, Wu \etal \cite{wu2023net} introduced a landmark-aware network to reduce expression ambiguity by leveraging facial landmarks to mitigate label noise.
Similarly, Lee \etal \cite{lee2023latent} proposed LatentOFER, a model that detects and restores occluded facial parts, improving recognition accuracy under occlusions.
To address the imbalanced FER problem, Zhang \etal \cite{zhang2024leave} developed a re-balanced attention map technique that regularizes the model, allowing it to extract transformation-invariant information from minority classes.

% In contrast, research on dynamic FER primarily aims to develop models capable of generating robust spatial-temporal representations of facial expressions. The introduction of the Transformer to DFER, as seen in Former-DFER \cite{zhao2021former}, marked a significant advancement, demonstrating superior performance compared to methods based on 3D CNNs or CNN-RNNs. Subsequent studies have further enhanced performance by addressing issues such as noisy frames \cite{li2022nr} and the intensity of varied frames \cite{li2023intensity}. Moreover, Wang \etal \cite{wang2023rethinking} introduced M3DFEL to tackle the challenge of imbalances between short- and long-term temporal relationships in DFER. The concept of soft labeling \cite{kawamura2024midas} has also been applied to reduce data ambiguity.

\textbf{Vision-language-based FER Models:} 
Unlike traditional FER models, vision-language-based models do not rely on explicit classifiers.
Instead, they compute final classification results by comparing the similarity between visual and textual features, selecting the highest similarity score as the prediction.
For example, Li \etal \cite{li2023cliper} introduced CLIPER, a unified framework for FER, which leverages multiple expression text descriptors to learn fine-grained expression representations.
% Zhao \etal \cite{zhao2023dferclip} proposed DFER-CLIP, which is designed for dynamic FER with temporal learning, introducing textual descriptions of facial behavior for contrastive learning.
Zhang \etal \cite{zhang2023weakly} presented the CLEF model, which conducts text-driven contrastive learning for facial behavior understanding, utilizing activity descriptions and coarse-grained information from specific datasets.
% However, the reliance on extensive labeled facial data in these models presents a significant challenge in data acquisition.
Despite these advancements, vision-language-based FER models face challenges related to the acquisition of extensive labeled facial data.
Foteinopoulou \etal \cite{foteinopoulou2023emoclip} introduces a novel vision-language model utilizing sample-level text descriptions as natural language supervision for facial expression video to enhance zero-shot FER classification.
However, labeling detailed facial action descriptions remains a challenge, requiring expertise in perceiving nuanced facial movements.
% Nonetheless, this approach encounters the difficulty of labeling detailed facial action descriptions, which requires expertise in perceiving nuanced facial movements.
The recent research by Li \etal \cite{li2024facial} introduced the FABA benchmark aimed at achieving facial expression analysis. 
Nevertheless, the au data and expression data within the FABA dataset are not interconnected, but rather generated independently, thereby lacking a coherent chain of thought.

\subsection{Facial Action Unit}
Facial Action Units (AUs), as defined by the Facial Action Coding System (FACS), serve as anatomically grounded descriptors that correspond to the contraction of distinct facial muscles or muscle groups. 
These AUs offer a standardized and objective framework for capturing subtle facial movements, including the formation of wrinkles, the elevation of eyebrows, and the depression of lip corners. 
This meticulous mapping enables researchers to dissect complex emotional expressions into their constituent AUs, facilitating a more profound comprehension of the interplay between facial muscle activity and the underlying emotional states.

Recent advancements in AU detection have introduced various deep learning-based methodologies. 
Some approaches segment the face into regions or patches~\cite{zhao2016deep,li2017eac,onal2019d} to learn AU-specific representations, while others explicitly model the interrelationships between AUs~\cite{li2019semantic,luo2022learning,yang2023fan}. 
More recent work has utilized vision transformers on RGB images~\cite{jacob2021facial} and multimodal data, incorporating RGB, depth, and thermal images~\cite{zhang2024multimodal}. 
Yin \etal~\cite{yin2024fg} used generative models combined with a pyramid CNN interpreter for AU detection, while Yang \etal~\cite{yang2023toward} jointly modeled AU-centered features, co-occurrences, and dynamics.

\subsection{Multi-modal Large Language Model}
Large Language Models (LLMs) exhibit remarkable reasoning capabilities in natural language processing tasks, prompting researchers to extend them to other modalities, such as images, audio, and motion, resulting in Multimodal Large Language Models (MLLMs). 
For instance, Flamingo~\cite{alayrac2022flamingo} integrates visual tokens into an LLM using cross-attention mechanisms, while BLIP-2~\cite{li2023blip} aligns visual features with text tokens within the LLM using the Q-Former model. 
Instruction tuning~\cite{wei2021finetuned} is commonly used to harmonize vision and language modalities, enhancing the capabilities of MLLMs.

Recent models like Mini-GPT4~\cite{zhu2023minigpt} and LLaVA~\cite{liu2023visual} utilize high-quality instruction tuning datasets to fine-tune MLLMs with minimal parameters. Instruct-BLIP~\cite{dai2023instructblip} introduces an instruction-aware visual feature extraction method, demonstrating promising zero-shot performance across multimodal tasks. 
mPlug-Owl~\cite{ye2023mplug} further fine-tunes both the visual encoder and abstractor during pre-training, while recent work by Xenos \etal~\cite{xenos2024vllms} leverages LLaVA's common sense knowledge for emotion recognition. 
AnyMAL~\cite{moon2023anymal} extends alignment beyond images to video, audio, and IMU sensor data, while Pink~\cite{Xuan_2024_CVPR} addresses the challenges of fine-grained image perception in MLLMs.
\subsection{Chain of Thought}

The Chain of Thought (CoT) mechanism has recently garnered significant attention in NLP due to its ability to improve model interpretability and performance~\cite{feng2024towards,wei2022chain,wang2022self}. 
By enabling models to generate step-by-step reasoning, CoT enhances tasks such as visual question answering and text generation, offering more coherent and accurate outputs~\cite{lyu2023faithful,mu2024embodiedgpt}.

Despite its success in NLP, the application of CoT in FER remains relatively unexplored. 
FER is inherently complex, involving the subtle interplay of facial action units and holistic expressions, which are challenging to decode and interpret. 
The introduction of CoT in FER could potentially offer a more intuitive and transparent analysis process, leading to improved model interpretability and accuracy.

Moreover, leveraging CoT in FER could unlock new possibilities in various applications, including mental health assessments, human-computer interaction, and even facial expression generation. 
By systematically analyzing the progression of emotions or micro-expressions, CoT could provide valuable insights into user states or generate more nuanced facial expressions in synthetic media.

% \section{Data Construct}

\begin{figure*}[tbh]
\begin{center}
\includegraphics[width=0.98\linewidth]{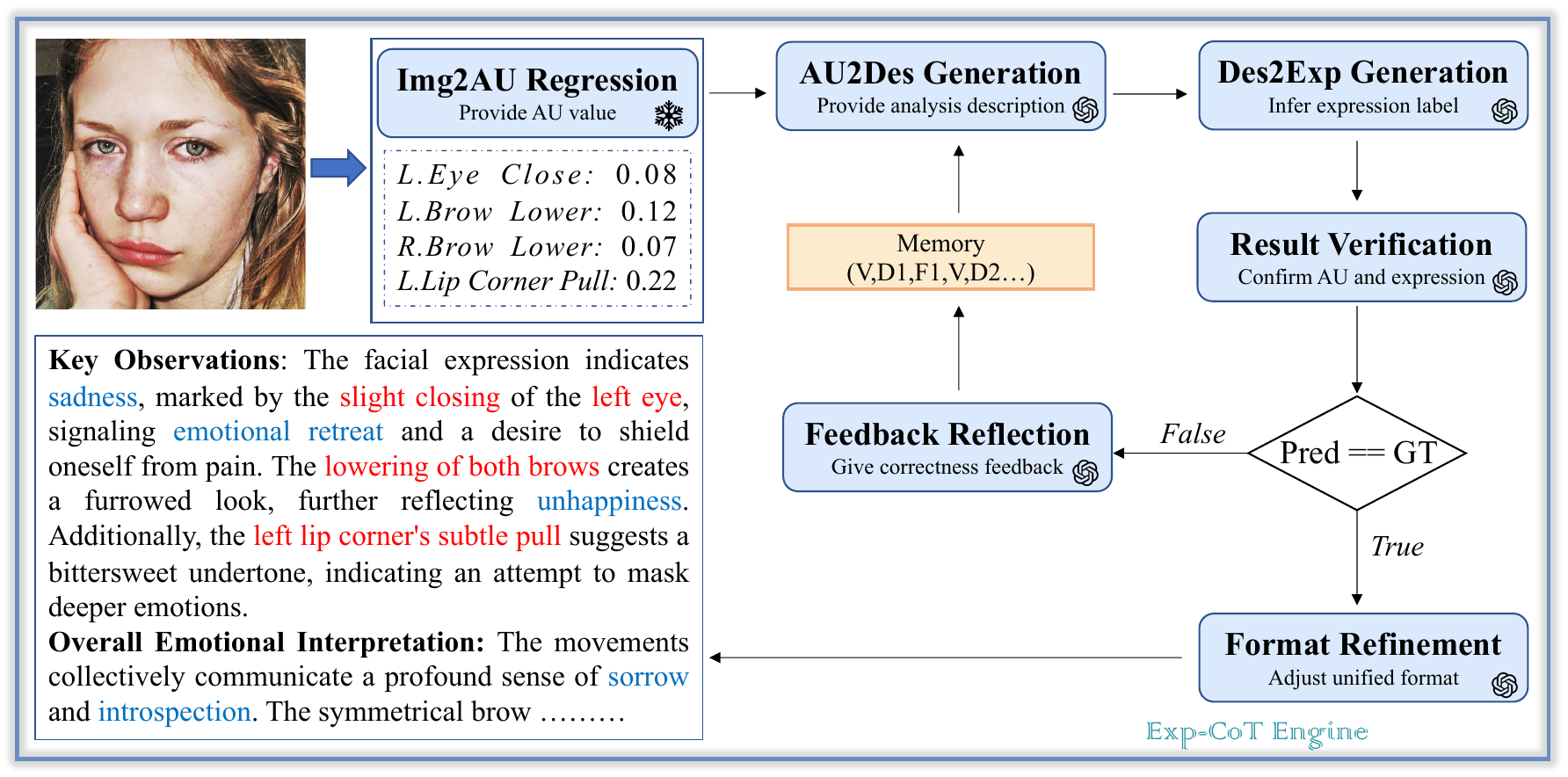}
\end{center}
\caption{Overview of the Exp-CoT Engine: This engine comprises six processes—Img2AU Regression, AU2Des Generation, Des2Exp Generation, Result Verification, Feedback Reflection, and Format Refinement. 
It utilizes advanced models, including the AU model and GPT-4o, to convert facial images into Action Units to generate detailed CoT of facial expressions.
The \textcolor{red}{red text} represent the AU name and its corresponding intensity, while the \textcolor{SkyBlue}{blue text} denote the potential emotions.
}
\label{fig:data engine}
\end{figure*}

\section{Methodology}
\label{sec:method}

\subsection{Overview}
Fig.\ref{fig: abstract} presents an overview of the proposed method, which is composed of two main components: Exp-CoT Engine and ExpLLM.
Our goal is to develop a model that generates a detailed chain of thought (CoT) for facial expressions, offering a comprehensive understanding of the underlying emotions.

In data construction phase, we design the Exp-CoT Engine to produce high-quality expression chain of thougt that are essential for training ExpLLM. 
This engine is pivotal in enhancing the model's ability to interpret and generate step-by-step descriptions of facial expressions, thereby improving its performance in facial expression recognition tasks.

As depicted in Figure \ref{fig:data engine}, the Exp-CoT Engine operates through a series of interconnected processes: 
Img2AU Regression, AU2Des Generation, Des2Exp Generation, Result Verification, Feedback Reflection, and Format Refinement. 
Each of these processes plays a critical role in ensuring the accuracy and reliability of the generated expression's CoT. 
The engine leverages state-of-the-art models, including an AU model and the latest GPT-4o, to perform tasks ranging from converting facial images to action units to generating comprehensive descriptions of facial expressions.

As illustrated in Figure \ref{fig:expllm}, ExpLLM is a multimodal model designed to analyze facial expressions through a sophisticated combination of multimodal processing and advanced language modeling techniques.
ExpLLM itself is built upon a multimodal model architecture similar to LLaVA, incorporating a visual encoder, a projector, and a decoder-only large language model. 
The model is fine-tuned using a parameter-efficient approach through LoRA, ensuring that it can effectively leverage the generated data pairs for improved performance in facial expression recognition tasks.
This architecture allows ExpLLM to process both visual and textual data, facilitating a deeper understanding of facial expressions. 

In summary, we provides a comprehensive framework for analyzing facial expressions, combining advanced data engineering techniques with sophisticated model architectures to achieve high accuracy and reasonable CoT in facial expression recognition. 
The Exp-CoT Engine and ExpLLM together represent a significant advancement in the field, offering a robust solution for interpreting and understanding human emotions through facial expressions.
We elaborate on the individual components of the proposed method in the following sections.

\subsection{Exp-CoT Engine}
% During the data construction phase, we develop a data engine termed the Exp-CoT Engine to create instruction-description data pairs. 
% The motivation behind developing the Exp-CoT Engine is to address the limitations of existing multimodal language models in accurately generating precise expression's CoT directly from facial images. 
% By utilizing a robust AU model and integrating GPT-4o in various stages, we aim to create a more accurate and reliable system for facial expression analysis. 
% The generated image-CoT, totaling 49,824 pairs from the RAF-DB and AffectNet datasets, serve as the training data for ExpLLM, enabling it to recognize and interpret facial expressions with enhanced precision.

% As illustrated in Fig.\ref{fig:data engine}, the Exp-CoT Engine encompasses six primary processes: Img2AU Regression, AU2Des Generation, Des2Exp Generation, Result Verification, Feedback Reflection, and Format Refinement.
% The Exp-CoT Engine leverages two fundamental models: the AU model and GPT-4o. Apart from the initial Img2AU process, the remaining five processes utilize GPT-4o, the latest iteration of the GPT model. 
% We also design a memory mechanism to store conversation history and feedback and input GPT-4o to refine analysis description.
% We elaborate on each process in detail in the subsequent sections.

During the data construction phase, we develop a specialized data engine, termed the Exp-CoT Engine, to generate instruction-description data pairs. 
The primary motivation behind developing the Exp-CoT Engine is to address the limitations of existing multimodal language models in accurately generating precise chains of thought (CoT) directly from facial images. 
By utilizing a robust AU model and integrating GPT-4o at various stages, our goal is to create a more accurate and reliable system for facial expression analysis.
The Exp-CoT Engine generates a total of 49,824 image-CoT pairs from the RAF-DB and AffectNet datasets. These pairs serve as the training data for ExpLLM, enabling it to recognize and interpret facial expressions with enhanced precision.

As illustrated in Figure \ref{fig:data engine}, the Exp-CoT Engine consists of six primary processes: Img2AU Regression, AU2Des Generation, Des2Exp Generation, Result Verification, Feedback Reflection, and Format Refinement. 
The Exp-CoT Engine leverages two fundamental models: the AU model and GPT-4o. While the initial Img2AU process relies on the AU model, the remaining five processes utilize GPT-4o, the latest iteration of the GPT model.
Additionally, we designed a memory mechanism to store conversation history and feedback, which is then used to refine analysis descriptions through GPT-4o. The following sections elaborate on each of these processes in detail.

\subsubsection{Img2AU Regression}

% Currently, MLLMs are unable to generate precise action units directly from facial images. 
% They can only produce coarse AU degrees, always have the problem of in missing or incorrectly ordered AU indices. 
% Given that existing AU models perform sufficient accuracy and stability for processing facial images in natural settings, we utilize the AU model to predict the AU sequence for each facial expression image.
% Specifically, we employ the state-of-the-art AU model from the FEAFA dataset \cite{yan2019feafa}, which offers 24 AUs \cite{gan2022feafa+} and continuous density values ranging from 0 to 1. 
% In comparison to the DISFA \cite{mavadati2013disfa} and BP4D \cite{zhang2014bp4d} datasets, the FEAFA dataset provides a more comprehensive set of AUs and finer-grained density annotations.
% The AU model generates a density value for each AU, enabling us to derive the AU sequence by sorting these density values. 
% Notably, we convert the AU indices to their corresponding names, such as transforming AU 1 to "Left Eye Close," to enhance human readability for subsequent interactions with Chat-GPT.

Current multimodal language models are limited in their ability to generate precise action units directly from facial images. 
These models often produce coarse AU predictions that suffer from missing or incorrectly ordered AU indices. 
To address this limitation, we utilize a specialized AU model that provides sufficient accuracy and stability for processing facial images in natural settings. 
This AU model predicts the AU sequence for each face image, which is essential for generating accurate facial expression chains of thought.
Specifically, we employ the state-of-the-art AU model from the FEAFA dataset \cite{yan2019feafa}, which offers 24 AUs \cite{gan2022feafa+} and continuous density values ranging from 0 to 1. 
Compared to the DISFA \cite{mavadati2013disfa} and BP4D \cite{zhang2014bp4d} datasets, the FEAFA dataset provides a more comprehensive set of AUs and finer-grained density annotations. 
% The AU model generates a density value for each AU, allowing us to derive the AU sequence by sorting these density values in descending order.

To enhance readability, we convert the AU indices into their corresponding names. 
For example, AU 1 is transformed into "Left Eye Close." 
This conversion improves the interpretability of the data and facilitates subsequent interactions with Chat-GPT, 
where these names are used to describe facial movements in a more intuitive manner.

\subsubsection{AU2Des Generation}

% After acquiring the AU sequence with corresponding intensities, we input this data into the GPT-4o model to generate a comprehensive description of the facial movement. 
% In addition to the positive AUs, we also include the negative AUs in the instructions. 
% It is crucial to highlight that we only provide AU data to the GPT-4o model, excluding the facial image itself. 
% This practice ensures that the expression is inferred solely from the perspective of AUs. 
% If the image is input to GPT, it would produce incorrect AU description and non-expression information.
% The instruction format is as follows:

% "Provided step-by-step analysis of the face based on the specified Facial Action Units. The positive AUs are \{'left eye close': 0.23, ...\}, and other AUs' density are 0. "

% We then get the analysis result from the GPT-4o model, which is a detailed description of the facial expression.
% However, the accuracy of the result is not guaranteed, necessitating subsequent verification.

After obtaining the AU sequence with corresponding intensities, we input this data into the GPT-4o model to generate a comprehensive description of the facial movements. 
To ensure a complete analysis, we include both positive AUs (those with non-zero intensities) and negative AUs (those with zero intensities) in the instructions.
It is important to note that only AU data is provided to the GPT-4o model, and the facial image itself is excluded. 
This approach ensures that the expression is inferred solely based on the AUs, thereby preventing the model from incorporating incorrect AU descriptions or irrelevant non-expression information that might arise if the facial image were included.
The instruction format provided to GPT-4o is as follows:

\textit{"Provided step-by-step analysis of the face based on the specified Facial Action Units. The positive AUs are \{'left eye close': 0.23, ...\}, and other AUs' density are 0."}

The output from GPT-4o is a detailed description of the facial expression based on the given AU data. 
However, due to potential limitations in the model's interpretation, the accuracy of this result is not guaranteed. 
Therefore, subsequent verification steps are necessary to ensure the reliability of the generated descriptions.

\subsubsection{Des2Exp Generation}
% To verify the correctness of the above description, we need first to generate the facial expression from the description.
% We also feed the description into the GPT-4o model to generate the facial expression.
% The instruction format is easy as follows:

% To ensure the accuracy of the facial expression description generated in the previous step, we need to validate it by generating the corresponding facial expression. 
This process involves feeding the description into the GPT-4o model, which then outputs the predicted facial expression.
The instruction format provided to GPT-4o is straightforward:

\textit{"Based on the description, just give me the expression result without other words. The result should only be one of the following: 'Surprise', 'Fear', 'Disgust', 'Happiness', 'Sadness', 'Anger', 'Neutral' and 'Contempt'."}

By limiting the response to these predefined categories, we ensure that the model's output is concise and relevant to the task of facial expression recognition.

\subsubsection{Result Verification}
% We need to verify the correctness of the generated expression from three perspectives: the positive AU sequence, the AU density, and the expression label. 
% We leverage GPT-4o to determine if the generated analysis description includes the positive AUs and whether the intensity described aligns with the AU density. 
% Additionally, we also consider if the generated expression label matches the ground truth label. 
% If the generated expression label is consistent with the ground truth, the expression-description pair is retained and proceeds to Format Refinement; otherwise, it proceeds to the feedback reflection process. 
% The instruction format is as follows:

To ensure the accuracy of the generated facial expression, we verify the results from three key perspectives: the positive AU sequence, the AU density, and the expression label. 
This verification process leverages GPT-4o to assess whether the generated analysis description includes the specified positive AUs and whether the intensities described align with the corresponding AU densities. 
Additionally, we check if the generated expression label matches the ground truth label.
The instruction format provided to GPT-4o is as follows:

\textit{"Determine whether the description contains AU \{left eye close: 0.23, ...\}, and the relevant degree corresponds to AU density. Also check if the generated expression label matches \{ground-truth expression label\}. 
Just give me the result 'Correct' or 'Incorrect', without any other words."}

This concise instruction ensures that the verification process is both efficient and accurate, allowing for reliable filtering of correct and incorrect expression-description pairs.
If the generated description is consistent with the ground truth, the expression-description pair is retained and proceeds to the Format Refinement stage. 
If not match, the pair is directed to the Feedback Reflection process for further refinement.

\subsubsection{Feedback Reflection}

% In instances where the generated expression label does not align with the ground truth, 
% it is imperative to provide feedback to the GPT-4o model to enhance the accuracy of the analysis. 
% The conversation history from this particular round should be stored in memory, and both the history and the feedback should be input into the GPT-4o model to refine the analysis description. 
When the generated description does not match the ground truth, it is essential to provide feedback to the GPT-4o model to improve the accuracy of its analysis. 
In such cases, the conversation history from the current round should be stored in memory. 
This history, along with the feedback, is then re-input into the GPT-4o model to refine the analysis description.
The feedback instruction should adhere to the following format:

\textit{"The just expression and analysis are incorrect. A revised response is requested based on the observations and analysis."}

% In scenarios where the round time exceeds a threshold, incorporating the correct expression label into the feedback instruction will supply the GPT-4o model with additional prior knowledge. 
% The instruction format for such cases should be:

In situations where the process exceeds a predefined time threshold, incorporating the correct expression label into the feedback can help supply GPT-4o with additional prior knowledge. 
For such cases, the instruction format should be:

\textit{"The generated expression and analysis remain incorrect. The correct expression is \{emotion\}. Please provide a revised analysis that aligns with this expression."}

This structured feedback mechanism ensures that the model can iteratively improve its outputs, enhancing the overall accuracy of the expression recognition process.

\subsubsection{Format Refinement}
% After multiple rounds of dialogue, the generated descriptions vary in length and structure, contain irrelevant sentences, and include a significant amount of noise.
% To address these issues and make model better understand the facial expression, we input the generated description into the GPT-4o model to refine the format and structure.
% The instruction format is as follows:

After multiple rounds of dialogue, the generated descriptions often vary in length and structure, contain irrelevant sentences, and include considerable noise. 
To address these issues and enhance the model's understanding of facial expressions, we input the generated descriptions into the GPT-4o model for refinement of format and structure.
The instruction format for this refinement is as follows:

\textit{"Enhance the expression description to make it more reasonable, presenting a logical flow of thought. 
Avoid the use of personal pronouns. 
New analysis should contain 3 parts: key observations, overall emotional interpretation, and conclusion.
The word count does not exceed 130.
Ensure that no AU indices (e.g., 'AU 1', 'AU 47') are included in the description. 
Original description: \{previous analysis\}."}

% It is worthnoting that we do not input AU index into the GPT-4o model, it cannot guarantee the AU index is correct corresponding to the AU name.
% Therefore, we need to remove the AU index from the description to make the description more reasonable.
% And we design the chain of thoughts by key observations, overall emotional interpretation, and conclusion to make the description more reasonable.
% Key observations contain the main AUs and their intensity, overall emotional interpretation should include all emotion and point out the main expression, and conclusion should include the final result.

It is important to note that we do not provide AU indices to the GPT-4o model, as it cannot reliably map the AU indices to their corresponding names. 
Therefore, removing AU indices from the description is essential for making the description more coherent and reasonable.
The structure of the refined description is designed to follow a clear chain of thought:
Key observations summarize the AU name, intensity and potential emotion; Overall emotional interpretation provide all emotions based on the AU sequence and point out the relationship among the emotions and movements; Conclusion concludes the final expression result.
This structured approach helps ensure that the final description is concise, logical, and effectively communicates the facial expression analysis.

We utilize the Exp-CoT Engine on the RAF-DB \cite{li2017reliable} and AffectNet \cite{mollahosseini2017affectnet} datasets to produce the instruction-description pairs essential for ExpLLM training. 
In total, we generate 49,824 instruction-description pairs, comprising 12,271 pairs from RAF-DB and 37,553 pairs from AffectNet. 
This generated data will be employed to train the ExpLLM model, enhancing its capabilities in facial expression recognition and enabling it to generate a chain of thought.

\subsubsection{Evaluation Metric}
\label{sec:exp-cot metric}
In terms of the expression CoT evaluation, it cannot be directly measured by nlp traditional metrics, such as BLEU~\cite{papineni2002bleu} or ROUGE~\cite{lin2004rouge}.
The reasons behind it contain three points:  
1. The sequence of Action Unit descriptions lacks consistency. For instance, descriptions may sometimes begin with the eyes and then proceed to the mouth, while at other times, the reverse order is used.
2. There is inconsistency in noun expressions, such as the use of "not raise" versus "drop" or "sadness" versus "sorrow.".
3. The internal structure and length of sentences vary significantly.
For example, consider the BLEU score for the following sentences: 

    \textit{"The small lowering of both brows conveys concern and distress."}

    \textit{"The left brow is slightly lowered, suggesting negative thoughts and distress, while the right brow mirrors this movement, enhancing the expression of sorrow."}

Despite their semantic similarity, the BLEU score~\cite{papineni2002bleu} for these sentences is only 0.005, which fails to accurately capture their resemblance. 
This highlights the limitations of traditional metrics like BLEU in evaluating the quality of expression CoTs.

To address this issue, we propose a novel evaluation metric, named Exp-CoT Score, to assess the quality of the generated expression CoTs.
This metric leverages GPT-4o evaluate each component of the expression CoT, including key observations, overall emotional interpretation, and conclusion, respectively.

Key Observations: GPT-4o assesses the similarity between the generated and ground truth descriptions by comparing the Action Unit names and their corresponding intensities. A score between 0 and 5 is assigned, with 5 indicating the highest level of agreement.

Overall Emotional Interpretation: GPT-4o evaluates the similarity of Action Unit combinations and their associated expressions (e.g., the combination of elevated eyelids and eyebrows indicating strong surprise, or subtle joyful expressions). This component is also scored between 0 and 5, with 5 representing the highest level of agreement.

Conclusion: GPT-4o checks the accuracy of the final expression label between the generated and ground truth descriptions. This component is scored as either 0 or 5, with 5 indicating a correct match.

Finally, the Exp-CoT Score is calculated by summing the three individual scores and normalizing the total. The final score ranges from 0 to 1, with a higher score indicating a more accurate and reasonable chain of thought.

\subsection{ExpLLM}

\begin{figure}[tbh]
\begin{center}
\includegraphics[width=\linewidth]{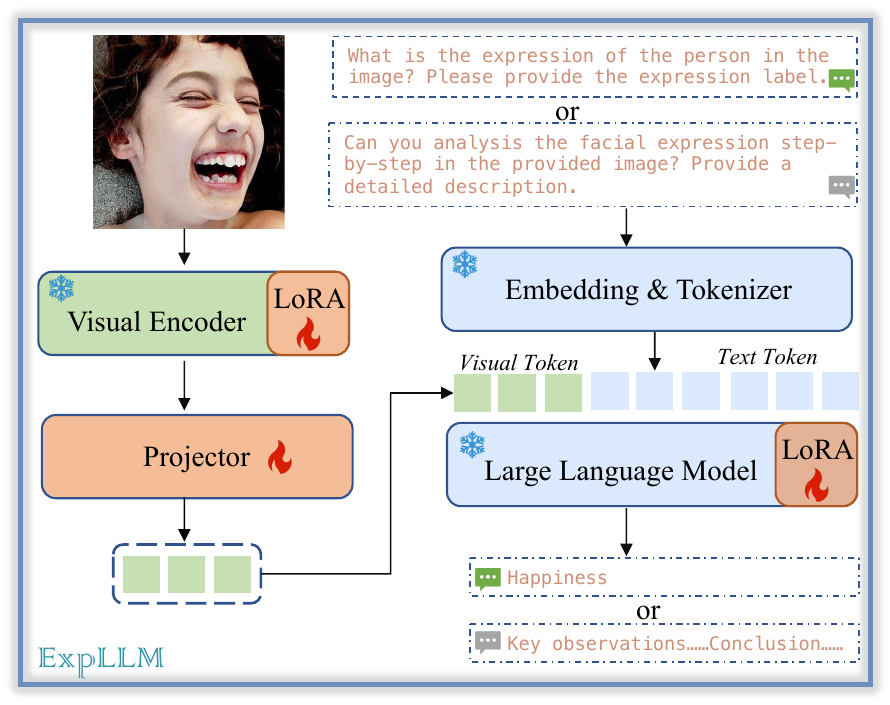}
\end{center}
\caption{A flowchart illustrating the ExpLLM methodology for analyzing facial expressions in images. 
The process begins with a visual encoder that utilizes LoRA to extract facial features from the image. 
The visual tokens are then projected and processed through an embedding and tokenizer before being fed into a large language model. 
The model can generate expression labels or detailed descriptions based on the analysis of the facial expression depicted in the image.
}
\label{fig:expllm}
\end{figure}

\subsubsection{Model Architecture}

As illustrated in Figure \ref{fig:expllm}, ExpLLM adopts an architecture similar to the multimodal model LLaVA~\cite{liu2023improvedllava,liu2023visual}, consisting of a visual encoder ($\Phi_{V}$), a projector ($\Phi_{P}$), and a decoder-only LLM ($\Phi_{L}$). 
Specifically, given an image $I$ and a user instruction sentence $S$, the visual encoder $\Phi_{V}$ transforms the image into a sequence of visual tokens $Z_V = \Phi_V(I)$. 
The instruction sentence $S$ is tokenized and embedded into a sequence of text tokens $Q_T$. 
The projector $\Phi_{P}$, comprising two linear layers, maps $Z_V$ into the input space of the LLM, resulting in $Z_T = \Phi_{P}(Z_V)$. 
Subsequently, $Z_T$ and $Q_T$ are concatenated and input into $\Phi_{L}$ to predict the subsequent words.

We employ the ViT-L/14 model as our visual encoder, utilizing weights pre-trained with DINOv2~\cite{oquab2023dinov2}. 
For the pre-trained language model, we use the instruction-tuned Vicuna-7B~\cite{vicuna2023}. 
Following the approach of LLaVA-1.5~\cite{liu2023improvedllava}, the projection layer consists of two fully connected layers.

Due to the extensive parameter size of the language model, fully fine-tuning the entire model is impractical given limited GPU resources. 
Additionally, fine-tuning both the language model and the visual encoder requires millions of image-text pairs to prevent model collapse, which is infeasible for the facial expression recognition (FER) task.
To enable efficient training while maximizing the benefits of multi-modal expression-analysis instruction conversations, we freeze the visual encoder and the language model, 
incorporating a limited set of trainable parameters within them. This approach mitigates semantic degradation in the visual encoder and LLM, which can result from limited instruction text data, and provides a parameter-efficient training strategy for facial expression recognition. 
Specifically, we employ LoRA~\cite{hu2021lora} to train ExpLLM.
The output feature \( h \) can be expressed as follows:
\begin{equation}
h = W_0 x + \Delta W x = (W_0 + \Delta W)x = W x + B A x,
\end{equation}
where \( W_0 \in \mathbb{R}^{d \times k} \) denotes the original weight matrix derived from the pre-trained model, and \( \Delta W \) signifies the learned weight perturbation introduced during the adaptation process. In this context, \( B \in \mathbb{R}^{d \times r} \) and \( A \in \mathbb{R}^{r \times k} \) represent the learnable low-rank matrices, with \( r \) being a significantly smaller rank than both \( d \) and \( k \). This formulation enables the incorporation of additional learnable parameters while maintaining the integrity and performance of the underlying pre-trained model. 

In our approach, we integrate LoRA modules into both the visual encoder and the LLM, fine-tuning them as well as the projector. 
This method results in a total of 8.7 million trainable parameters—a significantly smaller figure compared to full models and commonly used Convolutional Neural Networks or Vision Transformers. 
Unlike LLaVA, where the visual encoder is pre-trained on general image-text pairs, the visual tokens may insufficiently capture facial structures and expressions, as the encoder generally emphasizes broad semantic information over task-specific nuances.
Therefore, we augment the visual encoder with LoRA modules, specifically designed to enhance the representation of facial expression details within the visual tokens. 
This enhancement directs the model's focus towards more nuanced facial features, ultimately improving its performance in facial expression recognition tasks.

\subsubsection{Instruction Tuning}
We regard FER and CoT generation as two interconnected tasks and use multi-round conversation strategy to achieve multi-task learning.
In this approach, instruction-description and instruction-label data are structured as multi-round conversations, 
where CoT generation serves as an auxiliary task to enhance the accuracy of expression labels during training. 
This iterative conversation strategy enables the model to progressively refine its understanding of facial expressions, thereby improving its performance in FER tasks.
To enhance readability, expression labels—originally represented as numerical values corresponding to their respective classes—are converted into their corresponding expression names. The entire training process is conducted under text supervision.

ExpLLM undergoes a two-stage training process. 
In the first stage, we align the image and text by exclusively fine-tuning the projection layer on image-text pairs from CC3M \cite{sharma2018conceptual}. 
In the second stage, we freeze the visual encoder and LLM, and fine-tune the newly added LoRA module and projector on the expression instruction conversations. 
Consequently, ExpLLM can leverage the multi-modal conversation framework to perform facial expression recognition and generate a chain of thought with high accuracy. 

\section{Experiments}
\label{sec:experiments}
\subsection{Experimental Settings}
\subsubsection{Datasets}

\textbf{RAF-DB} \cite{li2017reliable} comprises an extensive collection of 15,000 facial images annotated with seven fundamental expressions: neutral, happiness, surprise, sadness, anger, disgust, and fear. 
Following prior works~\cite{zhang2023weakly,she2021dive,gao2024self}, we allocate 12,271 images for training, reserving the remaining 3,068 images for testing.

\textbf{AffectNet-8} \cite{mollahosseini2017affectnet} is the most extensive Facial Expression Recognition dataset, annotated with eight expressions: neutral, happiness, anger, sadness, fear, surprise, disgust, and contempt. 
This dataset contains a substantial 287,568 training samples and 4,000 testing samples. 
However, for practical reasons, we use a subset of 37,553 images (sourced from Kaggle) for training, as utilizing the entire training set is computationally intensive.

\textbf{AffectNet-7} \cite{mollahosseini2017affectnet} is derived from AffectNet-8 and includes seven expressions (neutral, happiness, anger, sadness, fear, surprise, and disgust), aligning with the RAF-DB expressions. 
To cross-evaluate the performance of our model trained on RAF-DB, we employ the AffectNet-7 test set, consisting of 3,500 images, as a benchmark.

\textbf{ExpW} \cite{zhang2015learning}, the Expression in-the-Wild Database, contains 91,793 faces downloaded using Google Image Search, with a confidence face level available for each image, allowing non-facial images to be filtered out. 
Each face image was manually annotated as one of the seven basic expression categories: angry, disgust, fear, happiness, sadness, surprise, and neutral, which is also consistent with the RAF-DB and AffectNet-7 datasets.
Due to the lack of a standard training/testing split, and following previous work \cite{arnaud2022thin}, we filtered images with a confidence face level greater than 60 and randomly selected 3,000 images for cross-evaluation.

\subsubsection{Implementation Details}
We utilize the ViT-L/14~\cite{dosovitskiy2020image} as our visual encoder, pre-trained with DINOv2~\cite{oquab2023dinov2} weights. 
For the pretrained Large Language Model, we employ the vicuna7B~\cite{vicuna2023}, an instruction-tuned model. 
Following the methodology of LLaVA-1.5, the projection layer consists of two fully connected layers. 
Additionally, LoRA modules are integrated into the query (q) and value (v) components of each self-attention layer within both the visual encoder and the LLM, configured with a hidden dimension \( r = 8 \).

All experiments were conducted utilizing eight NVIDIA 3090 GPUs, each equipped with 24GB of memory. 
The input images were processed at a resolution of 224$\times$224.
The RAF-DB is with original facial box, while others are with SCRFD~\cite{guo2021sample} detector to provide the facial box.
Since the SCRFD's results are similar to the RAF-DB's face box compared with other detectors, we can do cross-dataset evaluation directly.
Follows by previous works, we incorporated flip, rotation, erase, and color jittering as data augmentation techniques to enhance model robustness. 
For optimization across both stages, we employed AdamW as the optimizer.
In the first training stage, the projector was pretrained on the CC3M dataset for a single epoch, with a batch size of 256 and a weight decay set to 0.0. 
Following a warm-up period comprising 500 steps, the learning rate was initialized at 0.005 and subsequently decayed to 0 following a cosine schedule.
During the second stage, the model was trained on the facial expression recognition datasets in conjunction with CoT. 
Specifically, the training 60 epochs for RAF-DB and 40 epochs for AffectNet. 
This stage utilized a batch size of 128, achieved through gradient accumulation, and applied a weight decay of 0.05. 
The learning rate was initiated at 5e-4 after a substantial warm-up phase consisting of 20,000 steps.

\subsection{Evaluation Metrics}
We need to evaluate the performance of traditional facial expression recognition and the generated CoT.
Consistent with previous methods~\cite{zhao2021former,zhang2023weakly,gao2024self}, for the facial expression recognition, we adopt the standard metric, i.e. accuracy.
In terms of the generated CoT, we employ our Exp-CoT score as the evaluation metric, as described in Sec.\ref{sec:exp-cot metric}. 
This metric assesses the similarity of each constituent element (key observations, overall emotional interpretation, and conclusion) in the predicted CoT to the corresponding component in the ground truth CoT.

\subsection{Comparison on FER}

\subsubsection{Compared with State-of-the-Art Methods}
To demonstrate the efficiency of our proposed approach, we conduct a comparative analysis with state-of-the-art methods on the RAF-DB and AffectNet datasets. 
The effectiveness of our approach is measured in terms of accuracy, with the results presented in Table~\ref{tab:fer sota}. 
We fine-tune the ExpLLM model using the original dataset in conjunction with the expression CoT generated by the Exp-CoT Engine. 
The model attains an accuracy of 91.03\% on RAF-DB, markedly surpassing the current state-of-the-art benchmarks. 
Additionally, the model achieves an accuracy of 65.93\% on AffectNet-7 and 62.86\% on AffectNet-8, consistently outperforming existing approaches across all datasets.
\begin{table}[ht]
\centering
\caption{Performance comparison of state-of-the-art methods on FER datasets. *: zero-shot visual language model.}
\begin{tabular}{lcccc}
\hline
\rowcolor[HTML]{EFEFEF}
\textbf{Methods} & \textbf{Year} & \textbf{AffectNet-7} & \textbf{AffectNet-8} & \textbf{RAF-DB} \\ \hline
% RAN  & 59.50 & - & 86.90 \\ \hline
SCN~\cite{wang2020suppressing} & CVPR'20 & 63.40 & 60.23 & 87.03 \\ \hline
RUL~\cite{zhang2021relative} & NIPS'21 & 61.43 & - & 88.98 \\ \hline
DMUE~\cite{she2021dive} & CVPR'21 & - & 62.84 & 88.76 \\ \hline
VTFF~\cite{ma2021facial} & TAC'21 & 64.80 & 61.85 & 88.14 \\ \hline
EAC~\cite{zhang2022learn} & ECCV'22 & 65.32 & - & 89.99 \\ \hline
% FaRL\^{*} & CVPR'22 & 64.85 & - & 88.31 \\ \hline
Ada-DF~\cite{liu2023dual} & ICASSP'23 & 65.34 & - & 90.04 \\ \hline
PCL~\cite{liu2023pose} & CVPR'23 & 60.77 & - & 85.92 \\ \hline
CLEF~\cite{zhang2023weakly} & ICCV'23 & 65.66 & 62.77 & 90.09 \\ \hline
FRA~\cite{gao2024self} & CVPR'24 & 65.85 & - & 90.76 \\ \hline
% CAGE & CVPR'24 & 66.60 & 62.30 & - \\ \hline

% FaRL* \cite{} & CVPR'22 & 26.95 & 23.50 & 24.98 \\ \hline
% BLIP2* \cite{li2023blip} &  arxiv'23 & 32.86 & 28.53 & 43.78 \\ \hline
% Exp-CLIP* \cite{zhao2024enhancing} & arxiv'24 & 44.27 & 38.44 & 58.70 \\ \hline
\rowcolor[HTML]{EFEFEF}
ExpLLM & ours & \textbf{65.93} & \textbf{62.86} & \textbf{91.03} \\ \hline
\end{tabular}
\label{tab:fer sota}
\end{table}

\subsubsection{Cross-Dataset Evaluation}
We also conduct cross-dataset evaluation on the ExpW and AffectNet-7 datasets to assess the generalization capability of our model, which is trained on RAF-DB.
We have done the Synonym replacement, like sad and sadness, to align the expression labels across datasets.
For comparative analysis, we utilize publicly available models including DAN, Ada-DF, FaRL\cite{Zheng_2022_CVPR}, and Exp-CLIP\cite{zhao2024enhancing}. 
% We take the ExpLLM trained on RAF-DB, to cross evaluate on ExpW, and AffectNet-7. 
% with faces cropped using SCRFD\cite{guo2021sample}, as its results closely resemble those obtained using RAF-DB's face boxes. 
The comparison results are presented in Table~\ref{tab:cross_eval}.
It is evident that large vision-language models including FaRL, ExpCLip and our ExpLLM exhibit superior performance in cross-dataset evaluations compared to traditional models, 
underscoring their robust generalization ability. 
Notably, the ExpLLM achieves an accuracy of 68.13\% on ExpW and 51.86\% on AffectNet-7, significantly surpassing the FaRL and Exp-CLIP methods.
% Exp-CLIP &  arxiv'24 & 44.27 & 38.44 & 58.70 \\ \hline
% BLIP2 &  arxiv'24 & 45.27 & 39.44 & 59.70 \\ \hline
\begin{table}[h]
    \centering
    \caption{Comparison with methods on Cross evaluation. *: zero-shot visual language model. \dag : our reproduction on RAF-DB using the official codes.}
    \label{tab:performance_comparison}
    \begin{tabular}{lccc}
        \hline
        \rowcolor[HTML]{EFEFEF}
        \textbf{Methods} & \textbf{Year} & \textbf{ExpW} & \textbf{AffectNet-7} \\
        \hline
        DAN~\cite{wen2023distract} & Biomimetics'23 & 55.78 & 41.54 \\
        \hline
        Ada-DF~\cite{liu2023dual} & ICASSP'23 & 59.59 & 46.18 \\
        \hline
        FaRL\dag~\cite{Zheng_2022_CVPR} & CVPR'22 & 64.45 & 49.33 \\
        \hline
        Exp-CLip*~\cite{zhao2024enhancing} & arxiv'24 & - & 44.27 \\ \hline
        \rowcolor[HTML]{EFEFEF}

        \textbf{ExpLLM} & ours & \textbf{68.13} & \textbf{51.86} \\
        \hline 
        
    \end{tabular}
    \label{tab:cross_eval}
\end{table}

\begin{figure*}[th]
\begin{center}
\includegraphics[width=\linewidth]{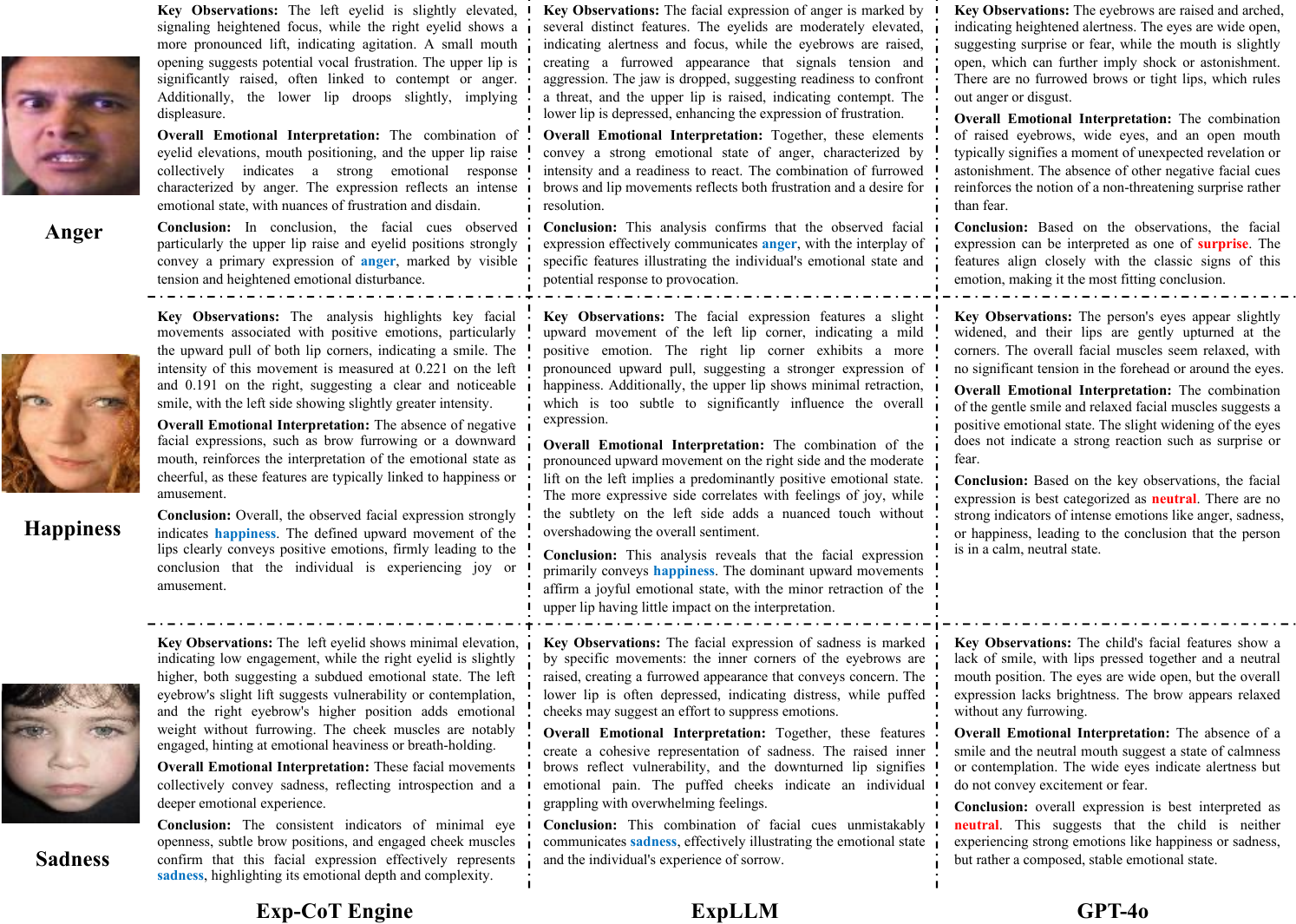}
\end{center}
\caption{Qualitative Comparison between ExpLLM and GPT-4o. First column: face image and ground truth expression label. Second column: CoT generated by Exp-CoT Engine. Third column: CoT generated by ExpLLM. Fourth column: CoT generated by GPT-4o.The \textcolor{SkyBlue}{blue expression state} is consistent with the ground truth, while the \textcolor{red}{red expression state} is incorrect.} 
\label{fig:compare samples}
\end{figure*}

\begin{figure*}[]
\begin{center}
\includegraphics[width=0.95\linewidth]{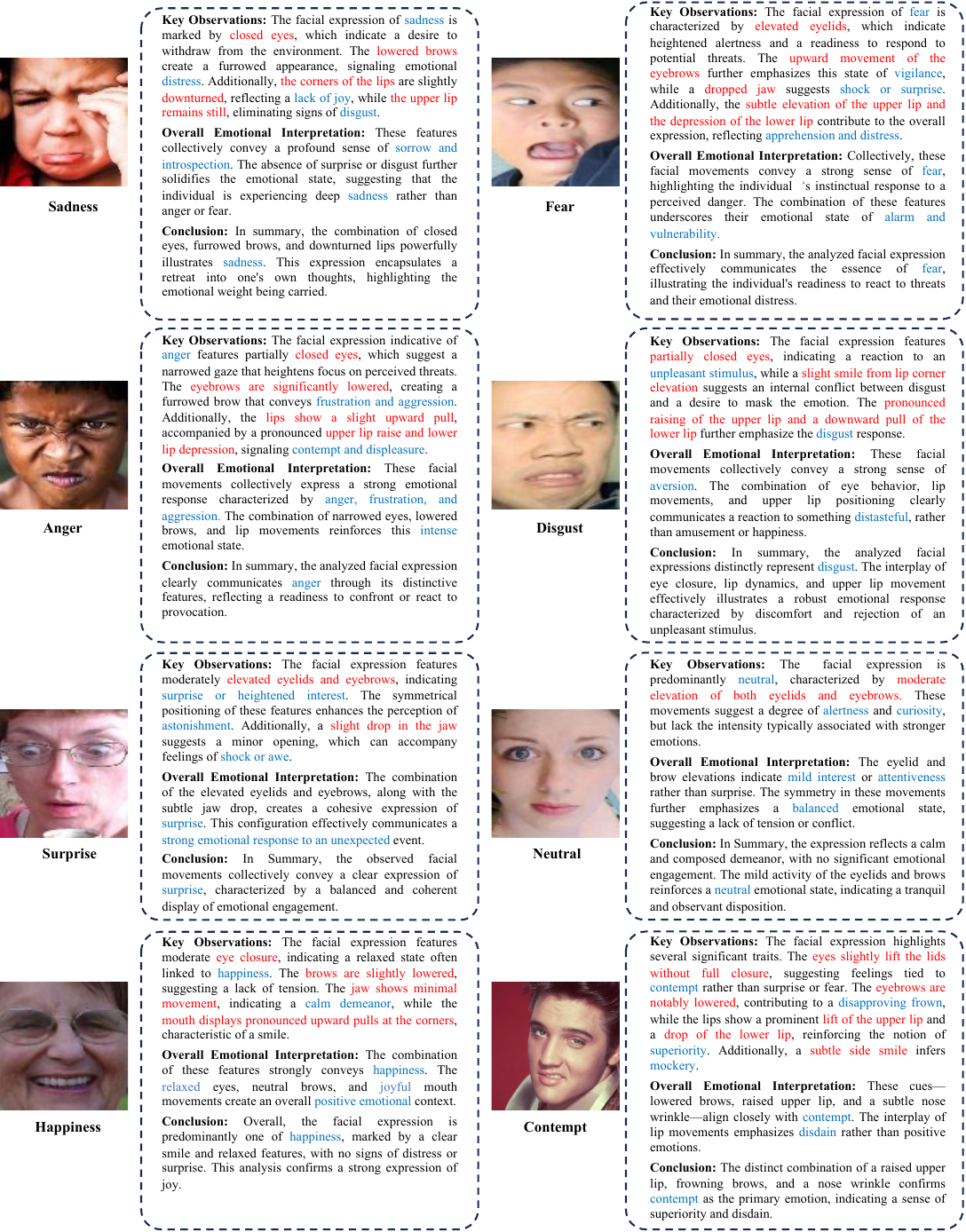}
\end{center}
\caption{Visualize the ExpLLM generated samples among eight different expressions.left is the face image and expression label, right is the generated CoT. The \textcolor{red}{red text} indicates the AU name and its corresponding intensity, while the \textcolor{SkyBlue}{blue text} represents the associated potential expression. 
}
\label{fig:samples}
\end{figure*}

\subsection{Comparsion on Expression CoT}
\subsubsection{Quantitative analysis}
We assess the quality of the generated CoT by comparing the predicted CoT with the ground truth CoT. 
For practical purposes, we randomly select 100 samples from the RAF-DB and AffectNet test sets and generate the ground-truth CoT for each sample using the Exp-CoT engine. 
Additionally, we utilize the ExpLLM model, which has been trained on the corresponding dataset, to generate the CoT for each sample. 
Similarly, we take GPT-4o with similar instruction to generate the CoT in the same format for each sample. 
Finally, we employ the Exp-CoT score to evaluate the quality of the generated CoT.

\begin{table}[h]
\centering
\caption{Comparison between GPT-4o and ExpLLM with Exp-Cot Score. KeyO.: Key Observations, Over.: Overall emotional interpretation, Conc.: Conclusion, ALL: Exp-CoT score.}
\label{tab:expcot score}
\begin{tabular}{@{}lcccc@{}}
    \hline
    Method & KeyO. & Over. & Conc. & ALL \\ \hline
    GPT-4o & 2.93 & 3.22 & 3.25 & 0.63 \\ 
    ExpLLM & \textbf{3.53} & \textbf{3.72} & \textbf{4.15} & \textbf{0.76} \\ \hline
\end{tabular}
\end{table}
As illustrated in Table~\ref{tab:expcot score}, by employing the Chain of Thought methodology, both models are capable of producing more sophisticated descriptions, ranging from key observations to conclusions, with a consistently enhanced accuracy.
ExpLLM attains a significantly higher Exp-CoT score compared to GPT-4o, particularly in the final conclusions. 
The lower score of key observations for GPT-4o indicates that it identifies fewer AU details, potentially overlooking some subtle yet crucial AU information. 
Consequently, the overall emotional interpretation of GPT-4o yields less nuanced emotion, attributed to the inadequacy of AU sequence information.
In contrast, ExpLLM can find more detailed AU information and generate a more comprehensive and accurate CoT, resulting in a higher Exp-CoT score across all components.

\subsubsection{Qualitative analysis}
We also conduct a comprehensive qualitative comparison between the CoT outputs generated by ExpLLM and GPT-4o, as illustrated in Fig.\ref{fig:compare samples}. 
The first column showcases the face image alongside the ground truth expression label, while the second column proposes the Exp-CoT generated according to the AU model and GPT-4o. 
The third column exhibits the CoT generated by ExpLLM, which is notably more precise and detailed compared to that of GPT-4o, displayed in the fourth column.
In detail, the key observations reveal that GPT-4o fails to capture micro-expression details, such as movements of the eyelids, jaw, and lips in the first sample, which are essential for accurate expression identification. 
Furthermore, some of GPT-4o's observations are incorrect. 
For instance, GPT-4o inaccurately states, "there are no furrowed brows or tight lips, which rules out anger or disgust," which contradicts the face image. 
These misinterpretations contribute to subsequent misrecognition.
In terms of overall emotional interpretation, ExpLLM also surpasses GPT-4o by providing a broader range of potential emotions based on the AU sequence and identifying the primary emotion and its relationship with other emotions. 
Due to GPT-4o's inability to capture micro-expression details, its generated expression labels lack accuracy, often mistakenly identifying expressions as neutral.

\subsection{Ablation Study}
\subsubsection{Training frequency between FER and CoT}

In our study, we observe a notable difference in the length of outputs generated by the two tasks. 
The CoT outputs are significantly longer, averaging around 130 words, while the FER results are typically limited to a single word.
Given the disparity in difficulty levels between two tasks, training them at the same frequency may compromise the accuracy of both.
As our primary objective remains expression recognition, an excessive emphasis on training the thought chain would inevitably compromise the accuracy of FER. 
Consequently, we aim to achieve an optimal balance between training both tasks concurrently.
\begin{table}[h]
\centering
\caption{Ablation Study on training rate about FER and CoT, *: evaluate the accuracy of the conclusion's expression.}
\label{tab:ablation rate}
\begin{tabular}{@{}lcc@{}}
    \hline
    FER : CoT & FER Acc & Exp-CoT Score \\ \hline
    1:0 & 89.71 & - \\
    0.8:0.2 & 90.79 & 0.75\\
    \rowcolor[HTML]{EFEFEF}
    0.75:0.25 & 91.03 & 0.78 \\
    0.67:0.33 & 90.54 & 0.79 \\
    0:1 & 80.08* & 0.83 \\ \hline
\end{tabular}
\end{table}

We conducted experiments on RAF-DB, and the results are presented in the Table \ref{tab:ablation rate}. 
The FER task is in one-round conversation, while the CoT task is still in two-round conversation with expression label.
Our findings indicate that when training is primarily based on FER, a moderate increase in CoT training enhances FER accuracy and provides greater explainability. 
At a 0.75:0.25 ratio, FER performance reaches its peak. 
However, further increasing the frequency of CoT training, despite yielding higher CoT scores, results in a decline in FER accuracy with diminishing returns in score improvement. 
Thus, we conclude that a training ratio of 0.75:0.25 represents the optimal balance for achieving the best outcomes in our study.

\subsubsection{Individual Training Components}
To assess the influence of each component of our training strategy on the overall performance, 
we conducted an ablation study using the RAF-DB dataset. 
The results, detailed in Table \ref{tab:each part}, encompass the following scenarios: 
training with pretraining (stage-1), training with a visual encoder (ViT), training with a large language model (LLM), and combined training with Chain of Thought (CoT).

\begin{table}[h]
    \centering
    \caption{Ablation study on Training with Different Combinations.}
    \begin{tabular}{cccc|c}
    \hline
    Stage-1 & ViT & LLM & CoT & FER Acc. (\%) \\
    % \hline
    \ding{55} & \ding{51} & \ding{51} & \ding{51} & 90.54 \\
    % \hline
    \ding{51} & \ding{55} & \ding{51} & \ding{51} & 82.09 \\
    % \hline
    \ding{51} & \ding{51} & \ding{55} & \ding{51} & 48.65 \\
    % \hline
    \ding{51} & \ding{51} & \ding{51} & \ding{55} & 89.71 \\
    % \hline
    \ding{51} & \ding{51} & \ding{51} & \ding{51} & 91.03 \\
    \hline
    \end{tabular}
    
    \label{tab:each part}
    \end{table}

Pretraining in Stage-1 proves advantageous for aligning visual and language features, a critical factor for optimal performance. 
Fine-tuning the LLM emerges as the most crucial element; without it, the model fails to achieve satisfactory results on the facial expression recognition task. 
Additionally, fine-tuning the ViT significantly enhances the learning of facial visual features, particularly subtle facial micro-expressions, resulting in an approximate 9\% improvement.

\subsection{Visual Samples}

We also present the predicted eight samples of the CoT generated by our model in Fig.\ref{fig:samples}, 
which encompass key observations, an overall emotional interpretation, and a conclusion. 
The key observations elucidate the intricate interplay between facial features and expression logic. 
The overall emotional interpretation offers a comprehensive analysis of all potential expressions, 
along with corresponding analyses and rationales. 
The conclusion provides a summative overview of the CoT, 
synthesizing insights from the overall emotional interpretation to elucidate the final emotional expression.

For instance, in the Fear sample, key observations include elevated eyelids, raised eyebrows, a dropped jaw, and varied lip movements. 
Additionally, it encompasses each AU's potential expression, such as vigilance, shock, apprehension, and distress. 
The overall emotional interpretation provides a deeper analysis considering the AU sequence; 
for example, these facial movements convey a strong sense of fear, 
and the combination of these features highlights their emotional state of alarm and vulnerability. 
The conclusion, based on the aforementioned analysis, identifies the final expression as fear and briefly assesses the character's status from a holistic perspective.

%  一个 AU 可能有很多种情感可能，但是一系列 AU 只会存在一种大方向的情感。需要把这些 AU 的逻辑和最终情感联系起来，得到初步的思维链。 进而分析都有哪些可能的表情，比如 （fear, alarm 和 vulnerability），给出合理的解释。 
% 最后基于解释的结果得出最终的表情结论，这个表情是 fear，因为这个表情是由 fear 和 alarm 组成的，而不是 vulnerability。

% 一个 AU 有很多可能情感， AU sequence 的相互影响能决定具体的情感表达，但是这种相互作用是什么呢？这个是需要解释的。

% overall emotional   提出了逻辑结构, 先铺垫了整体情感表达，然后递进式的分析了细节

% 基于整体描述，得出结论

\section{Conclusion}
\label{sec:conclusion}
In this paper, we introduce a novel method, ExpLLM, designed to offer an intuitive and transparent chain of thought for facial expressions. 
Initially, we develop an innovative data engine, the Exp-CoT Engine, which generates the analytical data employed in ExpLLM's instruction learning. 
We utilize the AU model and GPT-4o to transform facial images into AU to expression CoT.
Our approach models the expression CoT from three perspectives: key observations, overall emotional interpretation, and conclusion. 
Key observations encompass the AU name, its corresponding intensity, and the potential expression. 
The overall emotional interpretation provides an in-depth analysis by considering the AU sequence. 
The conclusion, derived from the preceding analysis, determines the final expression.
After pretraining and fine-tuning, ExpLLM can generate an accurate and reasonable CoT for facial expressions.
Extensive experiments on FER datasets reveal that ExpLLM surpasses current state-of-the-art methods in terms of accuracy.
And we also verify the feasibliity of the generated CoT in the three perspectives.
Future research will concentrate on CoT generation for video data, given its more complex action units and time sequence information.

\bibliographystyle{IEEEtran}
\bibliography{IEEEabrv, ref}

\begin{thebibliography}{10}
\providecommand{\url}[1]{#1}
\csname url@samestyle\endcsname
\providecommand{\newblock}{\relax}
\providecommand{\bibinfo}[2]{#2}
\providecommand{\BIBentrySTDinterwordspacing}{\spaceskip=0pt\relax}
\providecommand{\BIBentryALTinterwordstretchfactor}{4}
\providecommand{\BIBentryALTinterwordspacing}{\spaceskip=\fontdimen2\font plus
\BIBentryALTinterwordstretchfactor\fontdimen3\font minus
  \fontdimen4\font\relax}
\providecommand{\BIBforeignlanguage}[2]{{%
\expandafter\ifx\csname l@#1\endcsname\relax
\typeout{** WARNING: IEEEtran.bst: No hyphenation pattern has been}%
\typeout{** loaded for the language `#1'. Using the pattern for}%
\typeout{** the default language instead.}%
\else
\language=\csname l@#1\endcsname
\fi
#2}}
\providecommand{\BIBdecl}{\relax}
\BIBdecl

\bibitem{de2011facial}
F.~De~la Torre and J.~F. Cohn, ``Facial expression analysis,'' \emph{Visual
  analysis of humans: Looking at people}, pp. 377--409, 2011.

\bibitem{ekman1993facial}
P.~Ekman, ``Facial expression and emotion.'' \emph{American psychologist},
  vol.~48, no.~4, p. 384, 1993.

\bibitem{li2020deep}
S.~Li and W.~Deng, ``Deep facial expression recognition: A survey,'' \emph{IEEE
  transactions on affective computing}, vol.~13, no.~3, pp. 1195--1215, 2020.

\bibitem{bettadapura2012face}
V.~Bettadapura, ``Face expression recognition and analysis: the state of the
  art,'' \emph{arXiv preprint arXiv:1203.6722}, 2012.

\bibitem{shi2020human}
Y.~Shi, Z.~Zhang, K.~Huang, W.~Ma, and S.~Tu, ``Human-computer interaction
  based on face feature localization,'' \emph{Journal of Visual Communication
  and Image Representation}, vol.~70, p. 102740, 2020.

\bibitem{sivaram2019biometric}
M.~Sivaram, D.~Yuvaraj, G.~Megala, V.~Porkodi, M.~Kandasamy \emph{et~al.},
  ``Biometric security and performance metrics: Far, fer, cer, frr,'' in
  \emph{2019 International conference on computational intelligence and
  knowledge economy (ICCIKE)}.\hskip 1em plus 0.5em minus 0.4em\relax IEEE,
  2019, pp. 770--772.

\bibitem{lee2021development}
S.-C. Lee, G.-H. Lin, C.-C. Liu, E.-C. Chiu, and C.-L. Hsieh, ``Development of
  the cat--fer: A computerized adaptive test of facial emotion recognition for
  adults with schizophrenia,'' \emph{The American Journal of Occupational
  Therapy}, vol.~75, no.~1, pp. 7\,501\,205\,140p1--7\,501\,205\,140p11, 2021.

\bibitem{ekman1978facial}
P.~Ekman and W.~V. Friesen, ``Facial action coding system,''
  \emph{Environmental Psychology \& Nonverbal Behavior}, 1978.

\bibitem{zhi2020comprehensive}
R.~Zhi, M.~Liu, and D.~Zhang, ``A comprehensive survey on automatic facial
  action unit analysis,'' \emph{The Visual Computer}, vol.~36, no.~5, pp.
  1067--1093, 2020.

\bibitem{wang2024exploring}
Y.~Wang, W.~Chen, X.~Han, X.~Lin, H.~Zhao, Y.~Liu, B.~Zhai, J.~Yuan, Q.~You,
  and H.~Yang, ``Exploring the reasoning abilities of multimodal large language
  models (mllms): A comprehensive survey on emerging trends in multimodal
  reasoning,'' \emph{arXiv preprint arXiv:2401.06805}, 2024.

\bibitem{wang2024locllm}
D.~Wang, S.~Xuan, and S.~Zhang, ``Locllm: Exploiting generalizable human
  keypoint localization via large language model,'' in \emph{Proceedings of the
  IEEE/CVF Conference on Computer Vision and Pattern Recognition}, 2024, pp.
  614--623.

\bibitem{naeem2023i2mvformer}
M.~F. Naeem, M.~G. Z.~A. Khan, Y.~Xian, M.~Z. Afzal, D.~Stricker, L.~Van~Gool,
  and F.~Tombari, ``I2mvformer: Large language model generated multi-view
  document supervision for zero-shot image classification,'' in
  \emph{Proceedings of the IEEE/CVF Conference on Computer Vision and Pattern
  Recognition}, 2023, pp. 15\,169--15\,179.

\bibitem{li2017reliable}
S.~Li, W.~Deng, and J.~Du, ``Reliable crowdsourcing and deep
  locality-preserving learning for expression recognition in the wild,'' in
  \emph{Proceedings of the IEEE conference on computer vision and pattern
  recognition}, 2017, pp. 2852--2861.

\bibitem{mollahosseini2017affectnet}
A.~Mollahosseini, B.~Hasani, and M.~H. Mahoor, ``Affectnet: A database for
  facial expression, valence, and arousal computing in the wild,'' \emph{IEEE
  Transactions on Affective Computing}, vol.~10, no.~1, pp. 18--31, 2017.

\bibitem{zhang2015learning}
Z.~Zhang, P.~Luo, C.-C. Loy, and X.~Tang, ``Learning social relation traits
  from face images,'' in \emph{Proceedings of the IEEE international conference
  on computer vision}, 2015, pp. 3631--3639.

\bibitem{zhao2021learning}
Z.~Zhao, Q.~Liu, and S.~Wang, ``Learning deep global multi-scale and local
  attention features for facial expression recognition in the wild,''
  \emph{IEEE TIP}, vol.~30, pp. 6544--6556, 2021.

\bibitem{wang2019identity}
C.~Wang, S.~Wang, and G.~Liang, ``Identity-and pose-robust facial expression
  recognition through adversarial feature learning,'' in \emph{ACM MM}, 2019,
  pp. 238--246.

\bibitem{chen2021cross}
T.~Chen, T.~Pu, H.~Wu, Y.~Xie, L.~Liu, and L.~Lin, ``Cross-domain facial
  expression recognition: A unified evaluation benchmark and adversarial graph
  learning,'' \emph{IEEE TPAMI}, vol.~44, no.~12, pp. 9887--9903, 2021.

\bibitem{wang2020suppressing}
K.~Wang, X.~Peng, J.~Yang, S.~Lu, and Y.~Qiao, ``Suppressing uncertainties for
  large-scale facial expression recognition,'' in \emph{CVPR}, 2020, pp.
  6897--6906.

\bibitem{zhao2021robust}
Z.~Zhao, Q.~Liu, and F.~Zhou, ``Robust lightweight facial expression
  recognition network with label distribution training,'' in \emph{AAAI}, 2021,
  pp. 3510--3519.

\bibitem{wu2023net}
Z.~Wu and J.~Cui, ``La-net: Landmark-aware learning for reliable facial
  expression recognition under label noise,'' in \emph{ICCV}, 2023, pp.
  20\,698--20\,707.

\bibitem{lee2023latent}
I.~Lee, E.~Lee, and S.~B. Yoo, ``Latent-ofer: Detect, mask, and reconstruct
  with latent vectors for occluded facial expression recognition,'' in
  \emph{ICCV}, 2023, pp. 1536--1546.

\bibitem{zhang2024leave}
Y.~Zhang, Y.~Li, X.~Liu, W.~Deng \emph{et~al.}, ``Leave no stone unturned: Mine
  extra knowledge for imbalanced facial expression recognition,''
  \emph{NeurIPS}, vol.~36, 2024.

\bibitem{li2023cliper}
H.~Li, H.~Niu, Z.~Zhu, and F.~Zhao, ``Cliper: A unified vision-language
  framework for in-the-wild facial expression recognition,'' \emph{arXiv
  preprint arXiv:2303.00193}, 2023.

\bibitem{zhang2023weakly}
X.~Zhang, T.~Wang, X.~Li, H.~Yang, and L.~Yin, ``Weakly-supervised text-driven
  contrastive learning for facial behavior understanding,'' \emph{arXiv
  preprint arXiv:2304.00058}, 2023.

\bibitem{foteinopoulou2023emoclip}
N.~M. Foteinopoulou and I.~Patras, ``Emoclip: A vision-language method for
  zero-shot video facial expression recognition,'' in \emph{FG}, 2024.

\bibitem{li2024facial}
Y.~Li, A.~Dao, W.~Bao, Z.~Tan, T.~Chen, H.~Liu, and Y.~Kong, ``Facial affective
  behavior analysis with instruction tuning,'' \emph{arXiv preprint
  arXiv:2404.05052}, 2024.

\bibitem{zhao2016deep}
K.~Zhao, W.-S. Chu, and H.~Zhang, ``Deep region and multi-label learning for
  facial action unit detection,'' in \emph{Proceedings of the IEEE conference
  on computer vision and pattern recognition}, 2016, pp. 3391--3399.

\bibitem{li2017eac}
W.~Li, F.~Abtahi, Z.~Zhu, and L.~Yin, ``Eac-net: A region-based deep enhancing
  and cropping approach for facial action unit detection,'' in \emph{Face and
  Gesture Recognition}.\hskip 1em plus 0.5em minus 0.4em\relax IEEE, 2017, pp.
  103--110.

\bibitem{onal2019d}
I.~Onal~Ertugrul, L.~Yang, L.~A. Jeni, and J.~F. Cohn, ``D-pattnet: Dynamic
  patch-attentive deep network for action unit detection,'' \emph{Frontiers in
  computer science}, vol.~1, p.~11, 2019.

\bibitem{li2019semantic}
G.~Li, X.~Zhu, Y.~Zeng, Q.~Wang, and L.~Lin, ``Semantic relationships guided
  representation learning for facial action unit recognition,'' in
  \emph{Proceedings of the AAAI conference on artificial intelligence},
  vol.~33, 2019, pp. 8594--8601.

\bibitem{luo2022learning}
C.~Luo, S.~Song, W.~Xie, L.~Shen, and H.~Gunes, ``Learning multi-dimensional
  edge feature-based au relation graph for facial action unit recognition,''
  \emph{IJCAI}, 2022.

\bibitem{yang2023fan}
J.~Yang, J.~Shen, Y.~Lin, Y.~Hristov, and M.~Pantic, ``Fan-trans: Online
  knowledge distillation for facial action unit detection,'' in
  \emph{Proceedings of the IEEE/CVF Winter Conference on Applications of
  Computer Vision}, 2023, pp. 6019--6027.

\bibitem{jacob2021facial}
G.~M. Jacob and B.~Stenger, ``Facial action unit detection with transformers,''
  in \emph{CVPR}, 2021, pp. 7680--7689.

\bibitem{zhang2024multimodal}
X.~Zhang, H.~Yang, T.~Wang, X.~Li, and L.~Yin, ``Multimodal channel-mixing:
  Channel and spatial masked autoencoder on facial action unit detection,'' in
  \emph{WACV}, 2024, pp. 6077--6086.

\bibitem{yin2024fg}
Y.~Yin, D.~Chang, G.~Song, S.~Sang, T.~Zhi, J.~Liu, L.~Luo, and M.~Soleymani,
  ``Fg-net: Facial action unit detection with generalizable pyramidal
  features,'' in \emph{Proceedings of the IEEE/CVF Winter Conference on
  Applications of Computer Vision}, 2024, pp. 6099--6108.

\bibitem{yang2023toward}
J.~Yang, Y.~Hristov, J.~Shen, Y.~Lin, and M.~Pantic, ``Toward robust facial
  action units’ detection,'' \emph{Proceedings of the IEEE}, vol. 111,
  no.~10, pp. 1198--1214, 2023.

\bibitem{alayrac2022flamingo}
J.-B. Alayrac, J.~Donahue, P.~Luc, A.~Miech, I.~Barr, Y.~Hasson, K.~Lenc,
  A.~Mensch, K.~Millican, M.~Reynolds \emph{et~al.}, ``Flamingo: a visual
  language model for few-shot learning,'' \emph{NeurIPS}, vol.~35, pp.
  23\,716--23\,736, 2022.

\bibitem{li2023blip}
J.~Li, D.~Li, S.~Savarese, and S.~Hoi, ``Blip-2: Bootstrapping language-image
  pre-training with frozen image encoders and large language models,''
  \emph{arXiv preprint arXiv:2301.12597}, 2023.

\bibitem{wei2021finetuned}
J.~Wei, M.~Bosma, V.~Y. Zhao, K.~Guu, A.~W. Yu, B.~Lester, N.~Du, A.~M. Dai,
  and Q.~V. Le, ``Finetuned language models are zero-shot learners,''
  \emph{arXiv preprint arXiv:2109.01652}, 2021.

\bibitem{zhu2023minigpt}
D.~Zhu, J.~Chen, X.~Shen, X.~Li, and M.~Elhoseiny, ``Minigpt-4: Enhancing
  vision-language understanding with advanced large language models,''
  \emph{arXiv preprint arXiv:2304.10592}, 2023.

\bibitem{liu2023visual}
H.~Liu, C.~Li, Q.~Wu, and Y.~J. Lee, ``Visual instruction tuning,'' \emph{arXiv
  preprint arXiv:2304.08485}, 2023.

\bibitem{dai2023instructblip}
W.~Dai, J.~Li, D.~Li, A.~M.~H. Tiong, J.~Zhao, W.~Wang, B.~Li, P.~Fung, and
  S.~Hoi, ``Instructblip: Towards general-purpose vision-language models with
  instruction tuning,'' 2023.

\bibitem{ye2023mplug}
Q.~Ye, H.~Xu, G.~Xu, J.~Ye, M.~Yan, Y.~Zhou, J.~Wang, A.~Hu, P.~Shi, Y.~Shi
  \emph{et~al.}, ``mplug-owl: Modularization empowers large language models
  with multimodality,'' \emph{arXiv preprint arXiv:2304.14178}, 2023.

\bibitem{xenos2024vllms}
A.~Xenos, N.~M. Foteinopoulou, I.~Ntinou, I.~Patras, and G.~Tzimiropoulos,
  ``Vllms provide better context for emotion understanding through common sense
  reasoning,'' \emph{arXiv preprint arXiv:2404.07078}, 2024.

\bibitem{moon2023anymal}
S.~Moon, A.~Madotto, Z.~Lin, T.~Nagarajan, M.~Smith, S.~Jain, C.-F. Yeh,
  P.~Murugesan, P.~Heidari, Y.~Liu \emph{et~al.}, ``Anymal: An efficient and
  scalable any-modality augmented language model,'' \emph{arXiv preprint
  arXiv:2309.16058}, 2023.

\bibitem{Xuan_2024_CVPR}
S.~Xuan, Q.~Guo, M.~Yang, and S.~Zhang, ``Pink: Unveiling the power of
  referential comprehension for multi-modal llms,'' in \emph{Proceedings of the
  IEEE/CVF Conference on Computer Vision and Pattern Recognition (CVPR)}, June
  2024, pp. 13\,838--13\,848.

\bibitem{feng2024towards}
G.~Feng, B.~Zhang, Y.~Gu, H.~Ye, D.~He, and L.~Wang, ``Towards revealing the
  mystery behind chain of thought: a theoretical perspective,'' \emph{Advances
  in Neural Information Processing Systems}, vol.~36, 2024.

\bibitem{wei2022chain}
J.~Wei, X.~Wang, D.~Schuurmans, M.~Bosma, F.~Xia, E.~Chi, Q.~V. Le, D.~Zhou
  \emph{et~al.}, ``Chain-of-thought prompting elicits reasoning in large
  language models,'' \emph{Advances in neural information processing systems},
  vol.~35, pp. 24\,824--24\,837, 2022.

\bibitem{wang2022self}
X.~Wang, J.~Wei, D.~Schuurmans, Q.~Le, E.~Chi, S.~Narang, A.~Chowdhery, and
  D.~Zhou, ``Self-consistency improves chain of thought reasoning in language
  models,'' \emph{arXiv preprint arXiv:2203.11171}, 2022.

\bibitem{lyu2023faithful}
Q.~Lyu, S.~Havaldar, A.~Stein, L.~Zhang, D.~Rao, E.~Wong, M.~Apidianaki, and
  C.~Callison-Burch, ``Faithful chain-of-thought reasoning,'' \emph{arXiv
  preprint arXiv:2301.13379}, 2023.

\bibitem{mu2024embodiedgpt}
Y.~Mu, Q.~Zhang, M.~Hu, W.~Wang, M.~Ding, J.~Jin, B.~Wang, J.~Dai, Y.~Qiao, and
  P.~Luo, ``Embodiedgpt: Vision-language pre-training via embodied chain of
  thought,'' \emph{Advances in Neural Information Processing Systems}, vol.~36,
  2024.

\bibitem{yan2019feafa}
Y.~Yan, K.~Lu, J.~Xue, P.~Gao, and J.~Lyu, ``Feafa: A well-annotated dataset
  for facial expression analysis and 3d facial animation,'' in \emph{2019 IEEE
  International Conference on Multimedia \& Expo Workshops (ICMEW)}.\hskip 1em
  plus 0.5em minus 0.4em\relax IEEE, 2019, pp. 96--101.

\bibitem{gan2022feafa+}
W.~Gan, J.~Xue, K.~Lu, Y.~Yan, P.~Gao, and J.~Lyu, ``Feafa+: an extended
  well-annotated dataset for facial expression analysis and 3d facial
  animation,'' in \emph{Fourteenth International Conference on Digital Image
  Processing (ICDIP 2022)}, vol. 12342.\hskip 1em plus 0.5em minus 0.4em\relax
  SPIE, 2022, pp. 307--316.

\bibitem{mavadati2013disfa}
S.~M. Mavadati, M.~H. Mahoor, K.~Bartlett, P.~Trinh, and J.~F. Cohn, ``Disfa: A
  spontaneous facial action intensity database,'' \emph{IEEE Transactions on
  Affective Computing}, vol.~4, no.~2, pp. 151--160, 2013.

\bibitem{zhang2014bp4d}
X.~Zhang, L.~Yin, J.~F. Cohn, S.~Canavan, M.~Reale, A.~Horowitz, P.~Liu, and
  J.~M. Girard, ``Bp4d-spontaneous: a high-resolution spontaneous 3d dynamic
  facial expression database,'' \emph{Image and Vision Computing}, vol.~32,
  no.~10, pp. 692--706, 2014.

\bibitem{papineni2002bleu}
K.~Papineni, S.~Roukos, T.~Ward, and W.-J. Zhu, ``Bleu: a method for automatic
  evaluation of machine translation,'' in \emph{Proceedings of the 40th annual
  meeting of the Association for Computational Linguistics}, 2002, pp.
  311--318.

\bibitem{lin2004rouge}
C.-Y. Lin, ``Rouge: A package for automatic evaluation of summaries,'' in
  \emph{Text summarization branches out}, 2004, pp. 74--81.

\bibitem{liu2023improvedllava}
H.~Liu, C.~Li, Y.~Li, and Y.~J. Lee, ``Improved baselines with visual
  instruction tuning,'' 2023.

\bibitem{oquab2023dinov2}
M.~Oquab, T.~Darcet, T.~Moutakanni, H.~Vo, M.~Szafraniec, V.~Khalidov,
  P.~Fernandez, D.~Haziza, F.~Massa, A.~El-Nouby \emph{et~al.}, ``Dinov2:
  Learning robust visual features without supervision,'' \emph{arXiv preprint
  arXiv:2304.07193}, 2023.

\bibitem{vicuna2023}
\BIBentryALTinterwordspacing
W.-L. Chiang, Z.~Li, Z.~Lin, Y.~Sheng, Z.~Wu, H.~Zhang, L.~Zheng, S.~Zhuang,
  Y.~Zhuang, J.~E. Gonzalez, I.~Stoica, and E.~P. Xing, ``Vicuna: An
  open-source chatbot impressing gpt-4 with 90\%* chatgpt quality,'' March
  2023. [Online]. Available: \url{https://lmsys.org/blog/2023-03-30-vicuna/}
\BIBentrySTDinterwordspacing

\bibitem{hu2021lora}
E.~J. Hu, Y.~Shen, P.~Wallis, Z.~Allen-Zhu, Y.~Li, S.~Wang, L.~Wang, and
  W.~Chen, ``Lora: Low-rank adaptation of large language models,'' \emph{arXiv
  preprint arXiv:2106.09685}, 2021.

\bibitem{sharma2018conceptual}
P.~Sharma, N.~Ding, S.~Goodman, and R.~Soricut, ``Conceptual captions: A
  cleaned, hypernymed, image alt-text dataset for automatic image captioning,''
  in \emph{ACL}, 2018.

\bibitem{she2021dive}
J.~She, Y.~Hu, H.~Shi, J.~Wang, Q.~Shen, and T.~Mei, ``Dive into ambiguity:
  Latent distribution mining and pairwise uncertainty estimation for facial
  expression recognition,'' in \emph{Proceedings of the IEEE/CVF conference on
  computer vision and pattern recognition}, 2021, pp. 6248--6257.

\bibitem{gao2024self}
Z.~Gao and I.~Patras, ``Self-supervised facial representation learning with
  facial region awareness,'' in \emph{Proceedings of the IEEE/CVF Conference on
  Computer Vision and Pattern Recognition}, 2024, pp. 2081--2092.

\bibitem{arnaud2022thin}
E.~Arnaud, A.~Dapogny, and K.~Bailly, ``Thin: Throwable information networks
  and application for facial expression recognition in the wild,'' \emph{IEEE
  Transactions on Affective Computing}, vol.~14, no.~3, pp. 2336--2348, 2022.

\bibitem{dosovitskiy2020image}
A.~Dosovitskiy, L.~Beyer, A.~Kolesnikov, D.~Weissenborn, X.~Zhai,
  T.~Unterthiner, M.~Dehghani, M.~Minderer, G.~Heigold, S.~Gelly \emph{et~al.},
  ``An image is worth 16x16 words: Transformers for image recognition at
  scale,'' \emph{arXiv preprint arXiv:2010.11929}, 2020.

\bibitem{guo2021sample}
J.~Guo, J.~Deng, A.~Lattas, and S.~Zafeiriou, ``Sample and computation
  redistribution for efficient face detection,'' \emph{arXiv preprint
  arXiv:2105.04714}, 2021.

\bibitem{zhao2021former}
Z.~Zhao and Q.~Liu, ``Former-dfer: Dynamic facial expression recognition
  transformer,'' in \emph{ACM MM}, 2021, pp. 1553--1561.

\bibitem{zhang2021relative}
Y.~Zhang, C.~Wang, and W.~Deng, ``Relative uncertainty learning for facial
  expression recognition,'' \emph{Advances in Neural Information Processing
  Systems}, vol.~34, pp. 17\,616--17\,627, 2021.

\bibitem{ma2021facial}
F.~Ma, B.~Sun, and S.~Li, ``Facial expression recognition with visual
  transformers and attentional selective fusion,'' \emph{IEEE Transactions on
  Affective Computing}, vol.~14, no.~2, pp. 1236--1248, 2021.

\bibitem{zhang2022learn}
Y.~Zhang, C.~Wang, X.~Ling, and W.~Deng, ``Learn from all: Erasing attention
  consistency for noisy label facial expression recognition,'' in
  \emph{European Conference on Computer Vision}.\hskip 1em plus 0.5em minus
  0.4em\relax Springer, 2022, pp. 418--434.

\bibitem{liu2023dual}
S.~Liu, Y.~Xu, T.~Wan, and X.~Kui, ``A dual-branch adaptive distribution fusion
  framework for real-world facial expression recognition,'' in \emph{ICASSP
  2023-2023 IEEE International Conference on Acoustics, Speech and Signal
  Processing (ICASSP)}.\hskip 1em plus 0.5em minus 0.4em\relax IEEE, 2023, pp.
  1--5.

\bibitem{liu2023pose}
Y.~Liu, W.~Wang, Y.~Zhan, S.~Feng, K.~Liu, and Z.~Chen, ``Pose-disentangled
  contrastive learning for self-supervised facial representation,'' in
  \emph{Proceedings of the IEEE/CVF Conference on Computer Vision and Pattern
  Recognition}, 2023, pp. 9717--9728.

\bibitem{Zheng_2022_CVPR}
Y.~Zheng, H.~Yang, T.~Zhang, J.~Bao, D.~Chen, Y.~Huang, L.~Yuan, D.~Chen,
  M.~Zeng, and F.~Wen, ``General facial representation learning in a
  visual-linguistic manner,'' in \emph{Proceedings of the IEEE/CVF Conference
  on Computer Vision and Pattern Recognition (CVPR)}, June 2022, pp.
  18\,697--18\,709.

\bibitem{zhao2024enhancing}
Z.~Zhao, Y.~Cao, S.~Gong, and I.~Patras, ``Enhancing zero-shot facial
  expression recognition by llm knowledge transfer,'' \emph{arXiv preprint
  arXiv:2405.19100}, 2024.

\bibitem{wen2023distract}
Z.~Wen, W.~Lin, T.~Wang, and G.~Xu, ``Distract your attention: Multi-head cross
  attention network for facial expression recognition,'' \emph{Biomimetics},
  vol.~8, no.~2, p. 199, 2023.

\end{thebibliography}
% \newpage

% \section*{Biography Section}
% \input{texs/biography.tex}

% \vfill

\end{document}